\def\year{2018}\relax
\documentclass[letterpaper]{article} %DO NOT CHANGE THIS
\usepackage{aaai18}  %Required
\usepackage{times}  %Required
\usepackage{helvet}  %Required
\usepackage{courier}  %Required
\usepackage{url}  %Required
\usepackage{graphicx}  %Required
\usepackage{tree-dvips}
\usepackage{pgf,tikz}
\usepackage{tikz-qtree}
\usepackage{multirow}
\usepackage{amsmath}
\usepackage{mathtools}
\newcommand\newcite[1]{{\citeauthor{#1} (\citeyear{#1})}}
\newcommand\doc{{\it D }}
\newcommand\summary{{\it S }}
\newcommand\labels{{\it y }}
\newcommand\sentseq{{$(s_1, s_2, ..., s_n)$} }
\newcommand\capseq{{$(c_1, c_2, ..., c_p)$} }
\newcommand\labelseq{{$(y_1, y_2, ..., y_n)$} }
\title{Neural Extractive Summarization with Side Information}
\author{Shashi Narayan \quad Nikos Papasarantopoulos \quad Shay B. Cohen \quad Mirella
  Lapata \\Institute for Language, Cognition and Computation, School
  of Informatics, University of
  Edinburgh\\ \tt{\{shashi.narayan,nikos.papasa\}@ed.ac.uk} \tt{\{scohen,mlap\}@inf.ed.ac.uk}}

\begin{document}
\maketitle
\begin{abstract}
Most extractive summarization methods focus on the main body of the
document from which sentences need to be extracted. However, the gist
of the document may lie in side information, such as the title and
image captions which are often available for newswire articles. We
propose to explore side information in the context of single-document
extractive summarization. We develop a framework for single-document
summarization composed of a hierarchical document encoder and an
attention-based extractor with attention over side information. We
evaluate our model on a large scale news dataset. We show that
extractive summarization with side information consistently
outperforms its counterpart that does not use any side information, in
terms of both informativeness and fluency.
\end{abstract}

\section{Introduction}
\label{sec:intro}

Increased access to information and the massive growth in the global
news data have led to a growing demand from readers to spot emerging
trends, person mentions and the evolution of storylines in the
news \cite{liepins-EtAl:2017:EACLDemo}. The vast majority of this news
data contains textual documents, driving the need for automatic
document summarization systems aiming at acquiring key points in the
form of a short summary from one or more documents.

While it is not so challenging for humans to summarize text, automatic
summarization systems struggle with producing high quality
summaries. Both extractive and abstractive systems have been proposed
in recent years. Extractive summarization systems select sentences
from the document and assemble them together to often generate a
grammatical, fluent and semantically correct summary
\cite{jp-acl16,nallapati17,Yasunaga:2017:gcn}.
% \cite{krageback-cvsc14,Yin-ijcai15,nallapati-arxiv16,jp-acl16,nallapati17,Yasunaga:2017:gcn}.
Abstractive summarization systems, on the other hand, aim at building
an internal semantic representation and then generate a summary from
scratch
\cite{chenIjcai-16,nallapati-signll16,see-acl17,tanwan-acl17}. % ,paulus-socher-arxiv17}. 
Despite recent improvements, abstractive systems still struggle to
outperform extractive systems. This paper addresses the task of
single-document summarization and explores how we can further
improve the sentence selection process for extractive summarization.

Most extractive methods often focus on the main body of the document
from which sentences are extracted. Traditional methods manually
define features which are local in the context of each sentence or a
set of sentences which form the body of the document. Such features
include sentence position and length \cite{radev-lrec2004}, keywords
and the presence of proper nouns
\cite{Kupiec:1995binary,mani2001automatic}, 
% \cite{Kupiec:1995binary,mani2001automatic,SparckJones:2007}, 
frequency information such as content word frequency, composition
functions for estimating sentence importance from word frequency, and
the adjustment of frequency weights based on context \cite{nenkova-06}
and low-level event-based features describing relationships between
important actors in a document \cite{filatova-04event}.
%% Most extractive methods often focus on the main body of the document
%% from which sentences are extracted. Traditional methods manually
%% define features which are local in the context of each sentence or a
%% set of sentences which form the body of the
%% document \cite{Kupiec:1995binary,mani2001automatic,radev-lrec2004,filatova-04event,nenkova-06,SparckJones:2007}. 
Sentences are ranked for extraction based on the overlap with
features. Recent deep learning methods circumvent human-engineered
features using continuous sentence
features. \newcite{krageback-cvsc14} and \newcite{Yin-ijcai15} map
sentences to a continuous vector space which is used for similarity
measurement to reduce the redundancy in the generated
summaries. \newcite{jp-acl16} and \newcite{nallapati17} use recurrent
neural networks to read sequences of sentences to get a document
representation which they use to label each sentence for
extraction. These methods report state of the art results without
using any kind of linguistic annotation.

\begin{figure}[t!]
  \center{ % \footnotesize % scriptsize % footnotesize
    \begin{tabular}{| p{7.8cm} |}
      \hline 
      \vspace{0.02pt}
      \textbf{South Korean Prime Minister Lee Wan-koo offers to resign} \\ \hline
      \vspace{0.001pt}
      
      Seoul (CNN) South Korea's Prime Minister Lee Wan-koo offered to
      resign on Monday amid a growing political scandal. \\
      
      %% \vspace{0.1pt}
      
      Lee will stay in his official role until South Korean President
      Park Geun-hye accepts his resignation. He has transferred his
      role of chairing Cabinet meetings to the deputy prime minister
      for the time being, according to his office. \\
      
      %% \vspace{0.1pt}

      Park heard about the resignation and called it "regrettable,"
      according to the South Korean presidential office. \\
      
      %% \vspace{0.1pt}

      Calls for Lee to resign began after South Korean tycoon Sung
      Woan-jong was found hanging from a tree in Seoul in an apparent
      suicide on April 9. Sung, who was under investigation for fraud
      and bribery, left a note listing names and amounts of cash given
      to top officials, including those who work for the
      President. \\ 

      %% \vspace{0.1pt}

      Lee and seven other politicians with links to the South Korean
      President are under investigation. %% A special prosecutor's team
      %% has been established to investigate the case.
      cont... \\

      %% Lee had adamantly denied the allegations as the scandal
      %% escalated: "If there are any evidence, I will give out my
      %% life. As a Prime Minister, I will accept Prosecutor Office's
      %% investigation first."

      %% Park has said that she is taking the accusations very
      %% seriously. Before departing on her trip to Central and South
      %% America, she condemned political corruption in her country.

      %% "Corruption and deep-rooted evil are issues that can lead to
      %% taking away people's lives. We take this very seriously."

      %% "We must make sure to set straight this issue as a matter of
      %% political reform. I will not forgive anyone who is responsible
      %% for corruption or wrongdoing."

      %% Park is in Peru and is expected to arrive back to South Korea on
      %% April 27.

      %% CNN's Paula Hancocks contributed to this report.
      \hline
    \end{tabular} 
    
    \vspace{0.1pt}
    
    \begin{tabular}{| c l |}
      \hline 
      
      \vspace{0.1pt}
      
      \begin{minipage}{.1\textwidth}
        \includegraphics[width=0.75\linewidth]{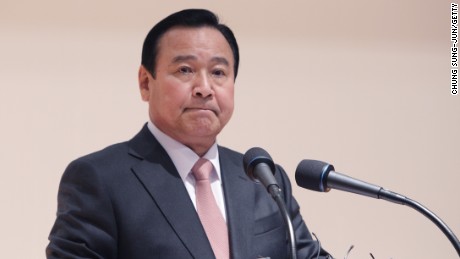}
      \end{minipage} & 
      \begin{minipage}{.31\textwidth}
      South Korean PM offers resignation over bribery scandal
      \end{minipage} \\

      \vspace{0.1pt}

      \begin{minipage}{.1\textwidth}
        \includegraphics[width=0.75\linewidth]{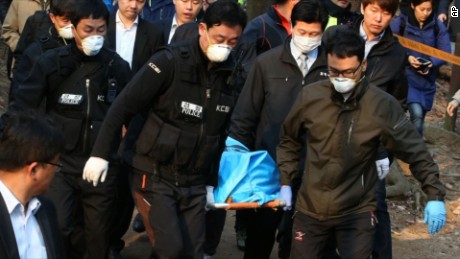}
      \end{minipage} & 
      \begin{minipage}{.31\textwidth}
        Suicide note leads to government bribery investigation
      \end{minipage} \\
      \hline             
    \end{tabular}
    
    \vspace{0.1pt}

    \begin{tabular}{| p{7.75cm} |}
      \hline 
      \vspace{0.1pt}
      \textbullet \hspace{0.1cm} Calls for Lee Wan-koo to resign began
      after South Korean tycoon Sung Woan-jong was found hanging from
      a tree in Seoul \\
      % \vspace{0.1pt}
      \textbullet \hspace{0.1cm} Sung, who was under investigation for
      fraud and bribery, left a note listing names and amounts of cash
      given to top officials \\
      \hline
    \end{tabular}
  }
  \caption{A CNN news article with story highlights and side
    information. The second block is the main body of the article. It
    comes with side information such as the title (first block) and
    the images with their captions (third block). The last block is
    the story highlights that assist in gathering information on the
    article quickly. These highlights are often used as the gold summary of
    the article in summarization literature.}\label{fig:cnnex}
\end{figure}

It is a challenging task to rely only on the main body of the document
for extraction cues, as it requires document understanding. Documents
in practice often have side information, such as the title, image
captions, videos, images and twitter handles, along with the main body
of the document. These types of side information are often available
for newswire articles.
%% Especially in the case of newswire documents, these are easily
%% available.
Figure \ref{fig:cnnex} shows an example of a newswire article taken
from CNN (\url{CNN.com}). It shows the side information such as the
title (first block) and the images with their captions (third block)
along with the main body of the document (second block). The last
block shows a manually written summary of the document in terms of
``highlights'' to allow readers to quickly gather information on
stories. As one can see in this example, gold highlights focus on
sentences from the fourth paragraph, i.e., on key events such as the
``PM's resignation'', ``bribery scandal and its investigation'',
``suicide'' and ``leaving an important note''. Interestingly, the
essence of the article is explicitly or implicitly mentioned in the
title and the image captions of the document.
%% side information. 
%% We propose to explore side information in the
%% context of single-document extractive summarization given the
%% availability of such additional information.

In this paper, we develop a general framework for single-document
summarization with side information. Our model includes a neural
network-based hierarchical document encoder and a hierarchical
attention-based sentence extractor. Our hierarchical document encoder
resembles the architectures proposed by \newcite{jp-acl16} % , \newcite{nallapati-arxiv16}
and \newcite{nallapati17}, in that it derives the document meaning
representation from its sentences and their constituent words. We also
use recurrent neural networks to read the sequence of sentences in the
document. Our novel sentence extractor combines this document meaning
representation with an attention mechanism \cite{bahdanau-arxiv14}
over the side information to select sentences of the input document as
the output summary. %% Our attention mechanism differs from both the
%% standard attention mechanism \cite{bahdanau-arxiv14} which is used to
%% locate the region of focus in the input, and the mechanism of
%% \newcite{jp-acl16} which directly extracts salient sentences after
%% reading them.  Instead, we use the attention mechanism to locate the
%% region of focus in the side information.

The idea of using additional information to improve extractive
summarization is less explored. Previous work has discussed the
importance of manually defined features using title words and words
with pragmatic cues (e.g., ``significant'', ``impossible'' and
``hardly'') for summarization. \newcite{Edmundson:1969} used a
subjectively weighted combination of these human-engineered features,
whereas
\newcite{Kupiec:1995binary} and \newcite{mani2001automatic} trained
their feature weights using a corpus. We explore the advantages of
side information in a neural network-based summarization
framework. Our proposed framework does not use any human-engineered
features and could exploit different types of side information. In
this paper, we conceptualize side information as the title of the
document and the image captions present in the document.\footnote{We
focus on textual side information. There are studies which show that
non-textual side information could be useful as well
(e.g., \citeauthor{hitschler-acl16} \citeyear{hitschler-acl16}) in
NLP. However, we leave non-textual side information for future work.}

%% In this work, we focus on textual side information to improve
%% extractive summarization.\footnote{There are studies which show that
%%   non-textual side information could be useful as well (e.g.,
%%   \cite{hitschler-acl16}) in NLP. However, we leave non-textual side
%%   information for future works.} We experiment with the title of the
%% document, image captions, the first sentence of the document and the
%% random sentences from the document. Note that the last two side
%% information come from the main body of the document itself. In our
%% experiment we want to explore if putting an extra attention at the
%% different part of the document itself might affect the sentence
%% selection process. Our experiments further explore the claim that the
%% first sentence of the document plays a crucial part in generating
%% summaries \cite{rush-acl15,nallapati-signll16}.

%% Our work differs from the trend of doing extractive summarization with
%% external knowledge from third party sources. \newcite{svore-emnlp07}
%% included features from news search query logs and Wikipedia entities
%% to summarize CNN articles. Recently, \newcite{li-coling16} used public
%% posts following a news article to improve automatic summary.  Our
%% model does not depend on any third party sources, it assumes that the
%% side information comes from the document itself. 

We evaluate our models automatically (in terms of ROUGE scores) on the
CNN news highlights dataset \cite{hermann-nips15}. Experimental
results show that our summarizer informed with side information
performs consistently better than the ones that do not use any side
information. We also conduct a human evaluation judging which type of
summary participants prefer. Our results overwhelmingly show that
human subjects find our summaries more informative and complete.

%% and by conducting human evaluations (in terms of which type of summary
%% participants prefer) on the CNN news highlights dataset
%% \cite{hermann-nips15}. Our results show that our summarizer informed
%% with side information performs consistently better than ones that do
%% not use any side information in generating informative and fluent
%% summaries.

%% We evaluate our models both automatically (in terms of ROUGE scores)
%% and with humans on two datasets: the CNN and DailyMail news highlights
%% corpora \cite{hermann-nips15}. Our results show that our summarizer
%% informed with side information performs consistently better than 
%%  which does not use any side information.

%% The rest of the paper is organized as follows. In \S\ref{sec:problem},
%% we formally describe the problem of extractive summarization with side
%% information. \S\ref{sec:model} describes our hierarchical
%% encoder-decoder model. \S\ref{sec:setup} describes our experimental
%% setup to assess our models. \S\ref{sec:results} presents our
%% results. \S\ref{sec:conclusion} concludes the paper.

\section{Problem Formulation}
\label{sec:problem}

In this section we formally define our extractive summarization
problem with side information. Given a document \doc consisting of a
sequence of sentences \sentseq and a sequence of pieces of side
information \capseq, we produce a summary \summary by selecting $m$
sentences from \doc (where $m < n$). We judge each sentence $s_i$ for
its relevance in the summary and label it with $y_i
\in \{0,1\}$ where $1$ indicates that $s_i$ should be considered for the
summary and $0$, otherwise. In this paper, we approach this problem in
a supervised setting where we aim to maximize the likelihood of the
set of labels \labels = \labelseq given the input document \doc and
model parameters $\theta$:

\begin{equation*}
p(y|\doc;\theta) = \prod_{i=1}^n p(y_i|\doc;\theta)
\end{equation*}

%% Here, we propose to do extractive summarization with captions. To do
%% this, we augment the document representation with captions: a document
%% \doc consists of a sequence of sentences \sentseq and a sequence of
%% captions \capseq. We aim at producing sequence of labels \labels =
%% \labelseq labelling each sentence $s_i$ in \doc.

The next section presents our model and discusses how it generates
summaries informed with side information.

\section{Summarization with Side Information}
\label{sec:model}

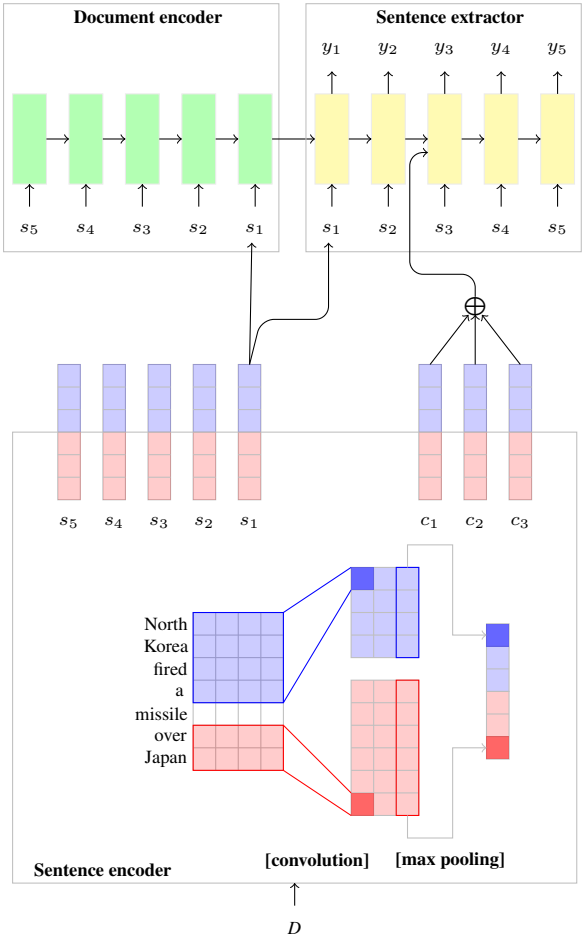
\begin{figure}[t!]
  \center{\scriptsize% footnotesize
    \begin{tikzpicture}[scale=0.6]
      \begin{scope}[shift={(-4,11)}] 
        % Outer shape
        \draw [lightgray] (-0.2,-1.5) rectangle (5.9, 4);
        \node at (3,3.7) {\textbf{Document encoder}};

        \draw [lightgray,fill=green,opacity=0.3] (0,0) rectangle (0.75,2);
        \draw [lightgray,fill=green,opacity=0.3] (1.25,0) rectangle (2,2);
        \draw [lightgray,fill=green,opacity=0.3] (2.50,0) rectangle (3.25,2);
        \draw [lightgray,fill=green,opacity=0.3] (3.75,0) rectangle (4.5,2);
        \draw [lightgray,fill=green,opacity=0.3] (5,0) rectangle (5.75,2);
        
        \draw [->] (0.375, -0.5) -- (0.375, 0) ;
        \node at (0.375,-1) {$s_5$};
        \draw [->] (1.625, -0.5) -- (1.625, 0) ;
        \node at (1.625,-1) {$s_4$};
        \draw [->] (2.875, -0.5) -- (2.875, 0) ;
        \node at (2.875,-1) {$s_3$};
        \draw [->] (4.125, -0.5) -- (4.125, 0) ;
        \node at (4.125,-1) {$s_2$};
        \draw [->] (5.375, -0.5) -- (5.375, 0) ;
        \node at (5.375,-1) {$s_1$};

        \draw [->] (0.75, 1) -- (1.25, 1) ;
        \draw [->] (2, 1) -- (2.50, 1) ;
        \draw [->] (3.25, 1) -- (3.75, 1) ;
        \draw [->] (4.5, 1) -- (5, 1) ;
        \draw [->] (5.75, 1) -- (6.7, 1) ;
      \end{scope}
      
      \begin{scope}[shift={(2.7,11)}]
        % Outer shape
        \draw [lightgray] (-0.2,-1.5) rectangle (5.9, 4);
        \node at (3,3.7) {\textbf{Sentence extractor}};

        \draw [lightgray,fill=yellow,opacity=0.3] (0,0) rectangle (0.75,2);
        \draw [lightgray,fill=yellow,opacity=0.3] (1.25,0) rectangle (2,2);
        \draw [lightgray,fill=yellow,opacity=0.3] (2.50,0) rectangle (3.25,2);
        \draw [lightgray,fill=yellow,opacity=0.3] (3.75,0) rectangle (4.5,2);
        \draw [lightgray,fill=yellow,opacity=0.3] (5,0) rectangle (5.75,2);
        
        \draw [->] (0.375, -0.5) -- (0.375, 0) ;
        \node at (0.375,-1) {$s_1$};
        \draw [->] (1.625, -0.5) -- (1.625, 0) ;
        \node at (1.625,-1) {$s_2$};
        \draw [->] (2.875, -0.5) -- (2.875, 0) ;
        \node at (2.875,-1) {$s_3$};
        \draw [->] (4.125, -0.5) -- (4.125, 0) ;
        \node at (4.125,-1) {$s_4$};
        \draw [->] (5.375, -0.5) -- (5.375, 0) ;
        \node at (5.375,-1) {$s_5$};

        \draw [->] (0.75, 1) -- (1.25, 1) ;
        \draw [->] (2, 1) -- (2.50, 1) ;
        \draw [->] (3.25, 1) -- (3.75, 1) ;
        \draw [->] (4.5, 1) -- (5, 1) ;
        
        \draw [->] (0.375, 2) -- (0.375, 2.50) ;
        \node at (0.375,3) {$y_1$};
        \draw [->] (1.625, 2) -- (1.625, 2.50) ;
        \node at (1.625,3) {$y_2$};
        \draw [->] (2.875, 2) -- (2.875, 2.50) ;
        \node at (2.875,3) {$y_3$};
        \draw [->] (4.125, 2) -- (4.125, 2.50) ;
        \node at (4.125,3) {$y_4$};
        \draw [->] (5.375, 2) -- (5.375, 2.50) ;
        \node at (5.375,3) {$y_5$};

      \end{scope}
      
      \begin{scope}[shift={(0,-2)}] 
        % Outer shape
        \draw [lightgray] (-4,-2.5) rectangle (8.5, 7.5);
        \node at (-2,-2.2) {\textbf{Sentence encoder}};
        \draw [->] (2.25, -3) -- (2.25, -2.5) ;
        \node at (2.25,-3.5) {\doc};

        % Outputs
        \draw [lightgray] (-3,6) rectangle (-2.5,9);
        \fill [red,opacity=0.2] (-3,6) rectangle (-2.5,7.5);
        \fill [blue,opacity=0.2] (-3,7.5) rectangle (-2.5,9);
        \draw [lightgray] (-3,6.5) -- (-2.5,6.5);
        \draw [lightgray] (-3,7) -- (-2.5,7);
        \draw [lightgray] (-3,7.5) -- (-2.5,7.5);
        \draw [lightgray] (-3,8) -- (-2.5,8);
        \draw [lightgray] (-3,8.5) -- (-2.5,8.5);
        \node at (-2.75,5.5) {$s_5$};
                
        \draw [lightgray] (-2,6) rectangle (-1.5,9);
        \fill [red,opacity=0.2] (-2,6) rectangle (-1.5,7.5);
        \fill [blue,opacity=0.2] (-2,7.5) rectangle (-1.5,9);
        \draw [lightgray] (-2,6.5) -- (-1.5,6.5);
        \draw [lightgray] (-2,7) -- (-1.5,7);
        \draw [lightgray] (-2,7.5) -- (-1.5,7.5);
        \draw [lightgray] (-2,8) -- (-1.5,8);
        \draw [lightgray] (-2,8.5) -- (-1.5,8.5);
        \node at (-1.75,5.5) {$s_4$};

        \draw [lightgray] (-1,6) rectangle (-0.5,9);
        \fill [red,opacity=0.2] (-1,6) rectangle (-0.5,7.5);
        \fill [blue,opacity=0.2] (-1,7.5) rectangle (-0.5,9);
        \draw [lightgray] (-1,6.5) -- (-0.5,6.5);
        \draw [lightgray] (-1,7) -- (-0.5,7);
        \draw [lightgray] (-1,7.5) -- (-0.5,7.5);
        \draw [lightgray] (-1,8) -- (-0.5,8);
        \draw [lightgray] (-1,8.5) -- (-0.5,8.5);
        \node at (-0.75,5.5) {$s_3$};

        \draw [lightgray] (0,6) rectangle (0.5,9);
        \fill [red,opacity=0.2] (0,6) rectangle (0.5,7.5);
        \fill [blue,opacity=0.2] (0,7.5) rectangle (0.5,9);
        \draw [lightgray] (0,6.5) -- (0.5,6.5);
        \draw [lightgray] (0,7) -- (0.5,7);
        \draw [lightgray] (0,7.5) -- (0.5,7.5);
        \draw [lightgray] (0,8) -- (0.5,8);
        \draw [lightgray] (0,8.5) -- (0.5,8.5);
        \node at (0.25,5.5) {$s_2$};

        \draw [lightgray] (1,6) rectangle (1.5,9);
        \fill [red,opacity=0.2] (1,6) rectangle (1.5,7.5);
        \fill [blue,opacity=0.2] (1,7.5) rectangle (1.5,9);
        \draw [lightgray] (1,6.5) -- (1.5,6.5);
        \draw [lightgray] (1,7) -- (1.5,7);
        \draw [lightgray] (1,7.5) -- (1.5,7.5);
        \draw [lightgray] (1,8) -- (1.5,8);
        \draw [lightgray] (1,8.5) -- (1.5,8.5);
        \node at (1.25,5.5) {$s_1$};

        \draw [rounded corners=5pt,->] (1.25, 9) -- (1.3, 11.7);
        \draw [rounded corners=5pt,->] (1.25, 9) -- (1.5, 10) -- (3, 10) -- (3, 11.7);
        
        \draw [lightgray] (5,6) rectangle (5.5,9);
        \fill [red,opacity=0.2] (5,6) rectangle (5.5,7.5);
        \fill [blue,opacity=0.2] (5,7.5) rectangle (5.5,9);
        \draw [lightgray] (5,6.5) -- (5.5,6.5);
        \draw [lightgray] (5,7) -- (5.5,7);
        \draw [lightgray] (5,7.5) -- (5.5,7.5);
        \draw [lightgray] (5,8) -- (5.5,8);
        \draw [lightgray] (5,8.5) -- (5.5,8.5);
        \node at (5.25,5.5) {$c_1$};

        \draw [lightgray] (6,6) rectangle (6.5,9);
        \fill [red,opacity=0.2] (6,6) rectangle (6.5,7.5);
        \fill [blue,opacity=0.2] (6,7.5) rectangle (6.5,9);
        \draw [lightgray] (6,6.5) -- (6.5,6.5);
        \draw [lightgray] (6,7) -- (6.5,7);
        \draw [lightgray] (6,7.5) -- (6.5,7.5);
        \draw [lightgray] (6,8) -- (6.5,8);
        \draw [lightgray] (6,8.5) -- (6.5,8.5);
        \node at (6.25,5.5) {$c_2$};
        
        \draw [lightgray] (7,6) rectangle (7.5,9);
        \fill [red,opacity=0.2] (7,6) rectangle (7.5,7.5);
        \fill [blue,opacity=0.2] (7,7.5) rectangle (7.5,9);
        \draw [lightgray] (7,6.5) -- (7.5,6.5);
        \draw [lightgray] (7,7) -- (7.5,7);
        \draw [lightgray] (7,7.5) -- (7.5,7.5);
        \draw [lightgray] (7,8) -- (7.5,8);
        \draw [lightgray] (7,8.5) -- (7.5,8.5);
        \node at (7.25,5.5) {$c_3$};
        
        \node at (6.25,10.3) {$\bigoplus$};
        \draw [->] (5.25,9) -- (6.1,10.1);
        \draw [->] (6.25,9) -- (6.25,10.1);
        \draw [->] (7.25,9) -- (6.4,10.1);
        
        \draw [rounded corners=5pt,->] (6.25,10.5) -- (6.25,11) --
        (4.8,11) -- (4.8,13.7) -- (5.2, 13.7) ; 
        
        % Convolutional mechanism
        \draw [lightgray] (0,0) -- (0, 3.5) -- (2,3.5) -- (2,0) -- (0,0);
        \draw [lightgray] (0,0.5) -- (2, 0.5); \draw [lightgray] (0,1) -- (2, 1);
        \draw [lightgray] (0,1.5) -- (2, 1.5); \draw [lightgray] (0,2) -- (2, 2);
        \draw [lightgray] (0,2.5) -- (2, 2.5); \draw [lightgray] (0,3) -- (2, 3); 
        \draw [lightgray] (0.5,0) -- (0.5, 3.5); \draw [lightgray] (1,0) -- (1, 3.5);
        \draw [lightgray] (1.5,0) -- (1.5, 3.5); 
        \node at (-0.6,3.25) {North};
        \node at (-0.6,2.75) {Korea};
        \node at (-0.5,2.25) {fired};
        \node at (-0.3,1.75) {a};
        \node at (-0.7,1.25) {missile};
        \node at (-0.5,0.75) {over};
        \node at (-0.6,0.25) {Japan};
        
        \fill [red,opacity=0.2] (0,0) rectangle (2,1);
        \draw [red] (0,0) rectangle (2,1);
        \fill [red,opacity=0.2] (3.5,-1) rectangle (5,2);
        \draw [lightgray] (3.5,-1) rectangle (5,2);
        \draw [lightgray] (3.5,-0.5) -- (5, -0.5); \draw [lightgray] (3.5,0) -- (5, 0);
        \draw [lightgray] (3.5, 0.5) -- (5, 0.5); \draw [lightgray] (3.5,1) -- (5, 1);
        \draw [lightgray] (3.5,1.5) -- (5, 1.5); 
        \draw [lightgray] (4,-1) -- (4, 2); \draw [lightgray] (4.5,-1) -- (4.5, 2);
        \fill [red,opacity=0.5] (3.5,-1) rectangle (4,-0.5);
        \draw [red] (2,0) -- (3.5,-1); \draw [red] (2,1) -- (3.5,-0.5); 
        
        \fill [blue,opacity=0.2] (0,1.5) rectangle (2,3.5);
        \draw [blue] (0,1.5) rectangle (2,3.5);
        \fill [blue,opacity=0.2] (3.5,2.5) rectangle (5,4.5);
        \draw [lightgray] (3.5,2.5) rectangle (5,4.5);
        \draw [lightgray] (3.5,3) -- (5, 3); \draw [lightgray] (3.5,3.5) -- (5, 3.5);
        \draw [lightgray] (3.5, 4) -- (5, 4); 
        \draw [lightgray] (4,2.5) -- (4, 4.5); \draw [lightgray] (4.5,2.5) -- (4.5, 4.5);
        \fill [blue,opacity=0.5] (3.5,4) rectangle (4,4.5);
        \draw [blue] (2,3.5) -- (3.5,4.5); \draw [blue] (2,1.5) -- (3.5,4); 
        
        \fill [blue,opacity=0.2] (6.5,3.25) rectangle (7,1.75);
        \draw [lightgray] (6.5,3.25) rectangle (7,1.75);
        \draw [lightgray] (6.5,2.75) -- (7, 2.75); \draw [lightgray] (6.5,2.25) -- (7, 2.25);
        \fill [blue,opacity=0.5] (6.5,2.75) rectangle (7,3.25);
        \fill [red,opacity=0.2] (6.5,1.75) rectangle (7,0.25);
        \draw [lightgray] (6.5,1.75) rectangle (7,0.25);
        \draw [lightgray] (6.5,1.25) -- (7, 1.25); \draw [lightgray] (6.5,0.75) -- (7, 0.75);
        \fill [red,opacity=0.5] (6.5,0.75) rectangle (7,0.25);
        
        \draw [blue] (4.5,2.5) rectangle (5,4.5);
        \draw [lightgray,->] (4.75,4.5) -- (4.75, 5) -- (5.75, 5) --
        (5.75, 3) -- (6.5, 3);
        \draw [red] (4.5,-1) rectangle (5,2);
        \draw [lightgray,->] (4.75,-1) -- (4.75, -1.5) -- (5.75, -1.5)
        -- (5.75, 0.5) -- (6.5, 0.5);
        
        \node at (2.75,-2) {\textbf{[convolution]}};
        \node at (5.7,-2) {\textbf{[max pooling]}};
      \end{scope}
      % The Muslim ban has backfired on Trump
      % Obama was the first african american president 
      % May God continue to bless the USA 
      \end{tikzpicture}
    } 
    \caption{Hierarchical encoder-decoder model for
    extractive summarization with side information. $s_1, \ldots, s_5$
    are sentences in the document and, $c_1$, $c_2$ and $c_3$
    represent side information.} \label{fig:architecture}
\end{figure}

Our extractive summarization framework consists of a hierarchical
encoder-decoder architecture assembled by recurrent neural networks
(RNNs) and convolutional neural networks (CNNs). The main components
of our model are a convolutional neural network sentence encoder, a
recurrent neural network document encoder and an attention-based
recurrent neural network sentence extractor. Our model exploits the
compositionality of the document. It reflects that a document is built
of a meaningful sequence of sentences and each sentence is built of a
meaningful sequence of words. With that in mind, we first obtain
continuous representations of sentences by applying single-layer
convolutional neural networks over sequences of word embeddings and
then we rely on a recurrent neural network to compose sequence of
sentences to get document embeddings. We model extractive
summarization as a sequence labelling problem using a standard
encoder-decoder architecture
\cite{sutskever-nips14}. First, the encoder reads the
sequence of sentences \sentseq in \doc and then, the decoder generates
a sequence of labels \labelseq labelling each sentence in \doc.
%% Our architecture resembles the sentence compression system of
%% \newcite{katja-emnlp15} where words in a sentence is labelled with
%% $1$ or $0$ to identify if they should be kept or deleted, and the
%% extractive summarization system of \newcite{jp-acl16} where the
%% decoder applies attention to directly extract salient sentences
%% after reading them. In contrast, our decoder uses an attention
%% mechanism \cite{bahdanau-arxiv14} to attend to side information to
%% improve extraction.
Figure~\ref{fig:architecture} presents the layout of our
model. In the following, we explain the main components of our model
in detail.

\subsection{Sentence Encoder}
\label{subsec:sentenc}

One core component of our hierarchical model is a convolutional
sentence encoder which encodes sentences (from the main body and the
side information) into continuous representations.\footnote{We tried
sentence/paragraph vector \cite{paravec} to infer sentence embeddings
in advance, but the results were inferior to those presented in this
paper with CNNs.}  CNNs \cite{leCun-1990} have shown to be very
effective in computer vision \cite{imagenet} and in
NLP \cite{nlpscratch}.
% various NLP
% tasks \cite{nlpscratch,kim-emnlp14,kalchbrenner-acl14,zhang-nips15,lei-emnlp15,kim-aaai16,jp-acl16}.
We chose CNNs in our framework for the following reasons. Firstly,
single-layer CNNs can be trained effectively and secondly, CNNs have
been shown to be effective in identifying salient patterns in the
input depending on the task. For example, for the caption generation
task \cite{showattendtell}, CNNs successfully identify salient objects
in the image for the corresponding words in the caption. We believe
that CNNs can similarly identify salient terms, e.g., named-entities
and events, in sentences that correlate with the gold summary. This
should in turn (i) optimize intermediate document representations in
both our document encoder and sentence extractor and (ii) assist the
attention mechanism to correlate salient information in the side
information and sentences, for extractive summarization.

% Some hierarchical model considers RNNs to encode sentences

Our model is a variant of the models presented in
\newcite{nlpscratch}, \newcite{kim-emnlp14} and \newcite{jp-acl16}. A
sentence $s$ of length $k$ in \doc is represented as a dense matrix
$W=[w_1, w_2, \ldots, w_k] \in R^{k \times d}$ where $w_i \in R^d$ is
the word embedding of the $i$th word in $s$. We apply a temporal
narrow convolution by using a kernel filter $K \in R^{h \times d}$ of
width $h$ for a window of $h$ words in $s$ to produce a new
feature. This filter is applied to each possible window of words in
$s$ to produce a feature map $f = [f_1, f_2, \ldots, f_{k-h+1}] \in
R^{k-h+1}$ where $f_i$ is defined:

\begin{equation*}
  f_i = \mathrm{ReLU}(K \circ W_{i:i+h-1}+b)
\end{equation*}

\noindent where, $\circ$ is the Hadamard Product followed by a sum
over all elements, $\mathrm{ReLU}$ is a rectified linear
activation\footnote{We use a smooth approximation to the rectifier,
i.e., the softplus function : $\mathrm{ReLU}(x)
= \mathrm{ln}(1+e^x)$.} and $b \in R$ is a bias term. We use the
$\mathrm{ReLU}$ activation function to accelerate the convergence of
stochastic gradient descent compared to sigmoid or tanh functions
\cite{imagenet}. We then apply max pooling over time \cite{nlpscratch}
over the feature map $f$ and get $f_{\mathrm{max}} = \mathrm{max}(f)$ as the
feature corresponding to this particular filter $K$. Max-pooling is
followed by local response normalization for better generalization
\cite{imagenet}. We use multiple kernels $K_h$ of width $h$ to compute
a list of features $f^{K_h}$. In addition, we use kernels of varying
widths to learn a set of feature lists $(f^{K_{h_1}}, f^{K_{h_2}},
\ldots)$. We concatenate all feature lists to get the final sentence
representation.\footnote{\newcite{jp-acl16} sum over feature lists to
  get the final sentence embedding. In contrast, we follow
  \newcite{kim-aaai16} and concatenate them. This seems to work best
  in our settings.}

The bottom part of Figure \ref{fig:architecture} briefly presents our
convolutional sentence encoder. Kernels of sizes $2$ (shown in red)
and $4$ (shown in blue) are applied 3 times each. The max pooling over
time operation leads to two feature lists $f^{K_{2}}$ and $f^{K_{4}}
\in R^3$. The final sentence embeddings have six dimensions. 
We use this sentence encoder to get sentence-level representations of
the sentences and side information (the title and image captions) of
the document \doc.

\subsection{Document Encoder}
\label{subsec:docenc}

The document encoder (shown in Figure \ref{fig:architecture}, top
left) composes a sequence of sentences to get a document
representation. The sentence extractor, along with attending the side
information, crucially exploits the document representation to
identify the local and global importance of a sentence in the document
to make a decision on whether it should be considered for the summary.

%% The document encoder (shown in Figure \ref{fig:architecture}, top
%% left) plays an integral role in this by composing a sequence of
%% sentences to get a document representation.

%% To make a decision on ``if a sentence should be considered for the
%% summary or not?'', our model conditions this decision on the document
%% to identify the local and global importance of the sentence. Document
%% encoder (shown in Figure \ref{fig:architecture}, top left) plays an
%% integral part in this by recursively composing sequence of sentences
%% to get a document representation which is later used in making those
%% informed decisions by sentence extractor.\footnote{Our encoder does a
%%   shallow analysis of the document at sentence level. It would be
%%   interesting to capture the signals from the syntactic or discourse
%%   analysis, e.g. Rhetorical Structure Theory (RST) \cite{rst} parsing,
%%   of the document. We don't pursue this direction here, we leave it
%%   for future work.}

We use a recurrent neural network with Long Short-Term Memory (LSTM)
cells to avoid the vanishing gradient problem when training long
sequences \cite{lstm}. Given document \doc consisting of a sequence of
sentences $(s_1, s_2, ..., s_n)$, we follow a common practice and feed
sentences in reverse order
\cite{sutskever-nips14,lijurafsky-acl15,katja-emnlp15}. This way we
make sure that the network does not omit top sentences of the document
which are particularly important for summarization
\cite{rush-acl15,nallapati-signll16}. At time step $t$, the hidden
state $h_t = \mathrm{LSTM}(s_{n-t+t}, h_{t-1})$ is updated as:

\begin{align*}
  \left[
  \begin{tabular}{c} $f_t$ \\ $i_t$ \\ $o_t$ \\ $\tilde{c}_t$ \end{tabular}
  \right] &= \left[ \begin{tabular}{c} $\sigma$ \\ $\sigma$
      \\ $\sigma$ \\ $\mathrm{tanh}$ \end{tabular} \right] W \cdot
  \left[ \begin{tabular}{c} $h_{t-1}$ \\ $s_{n-t+1}$ \end{tabular}
    \right] \\ c_t &= f_t \odot c_{t-1} + i_t \odot \tilde{c}_t \\ h_t
  & = o_t \odot \mathrm{tanh}(c_t)
\end{align*}

\noindent where the operator $\odot$ denotes element-wise
multiplication and $W$ are the learned parameters of the model.

\subsection{Sentence Extractor}
\label{subsec:sentext}

Our sentence extractor (Figure \ref{fig:architecture}, top right)
labels each sentence in the document with labels $1$ or $0$ by
implicitly estimating its relevance in the document and by directly
attending to the side information for importance cues. It is
implemented with another recurrent neural network with LSTM cells and
an attention mechanism \cite{bahdanau-arxiv14}. Our attention
mechanism differs from the standard practice of attending intermediate
states of the input (encoder). Instead, our extractor attends to the
side information in the document for cues. Given a document $\mbox{\it
D}:\langle (s_1, s_2, ..., s_n), (c_1, c_2, ..., c_p)\rangle$, it
reads sentences \sentseq in order and labels them one by one while
attending the side information \capseq consisting of the title and
image captions.
%% It also reads the side information in
%% order as they appear in the document. For example, if we consider the
%% title and the image captions as the side information, $c_1$ will be
%% the title and $c_2, \ldots, c_p$ will be the image captions in order
%% as they appear in the document. 
Given sentence $s_t$ at time step $t$, it returns a probability
distribution over labels as:

\begin{align*}
  p(y_t|s_t,D) &= \mathrm{softmax}(g(h_t,h_t'))\\
  g(h_t,h_t') &= U_o(V_h h_t + W_h' h_t') \\
  h_t &= \mathrm{LSTM}(s_t, h_{t-1}) \\
  h_t' &= \sum_{i=1}^{p} \alpha_{(t,i)} c_i, \\
  \mbox{where } \alpha_{(t,i)} &= \frac{\exp(h_t c_i)}{\sum_j \exp(h_t c_j)}
\end{align*}

\noindent where $g(\cdot)$ is a single-layer neural network with
parameters $U_o$, $V_h$ and $W_h'$. $h_t$ is an intermediate RNN state
at time step $t$. The dynamic context vector $h_t'$ is essentially the
weighted sum of the side information in the document. Figure
\ref{fig:architecture} summarizes our model. For each labelling
decision, our network considers the encoded document meaning
representation, sentences labeled so far and the side information.

\subsection{Summary Generation}

We rank sentences in the document \doc by $p(1|s_t,D,\theta)$, the
confidence scores assigned by the softmax layer of the sentence
extractor and generate a summary $S$ by assembling together the $m$
best ranked sentences.

\section{Experimental Setup}
\label{sec:setup}

This section presents our experimental setup for the assessment of our
models. We discuss the training and the evaluation dataset. We also
explain how we augment existing datasets with side information and
describe implementation details to facilitate the replication of our
results. We present a brief description of our baseline systems.

%% and of the metrics we use for human and automatic evaluation.

\subsection{Training and Test data}

To train our model, we need documents annotated with sentence
importance information, i.e., each sentence in a document is labelled
with 1 (summary-worthy) or 0 (not summary-worthy). For our purposes,
we used an augmented version of the CNN dataset
\cite{hermann-nips15}.\footnote{\newcite{hermann-nips15} have also
  released the DailyMail dataset, but we do not report our results on
  this dataset. We found that the script written
  by \citeauthor{hermann-nips15} to crawl DailyMail articles
  mistakenly extracts image captions as part of the main body of the
  document. As image captions often don't have sentence boundaries,
  they blend with the sentences of the document unnoticeably. This
  leads to the production of erroneous summaries.}

Our dataset is an evolved version of the CNN dataset first collected
by \newcite{svore-emnlp07} for highlight
generation. \newcite{svore-emnlp07} noticed that CNN articles often
come with ``story highlights'' to allow readers to quickly gather
information on stories. They collected a small dataset for evaluation
purposes. \newcite{woodsend-acl10} improved on this by collecting
9,000 articles and manually annotating them for sentence
extraction. Recently, \newcite{hermann-nips15} crawled 93K CNN
articles to build a large-scale corpus to set a benchmark for deep
learning methods. Since then, this dataset has been used for
single-document summarization
\cite{nallapati-signll16,jp-acl16,nallapati17,see-acl17,tanwan-acl17}.
\newcite{jp-acl16} annotated this dataset with the \newcite{woodsend-acl10} 
style gold annotation using a rule-based method judging each sentence
for its semantic correspondence to the gold
summary. \newcite{nallapati17} automatically extracted ground truth
labels such that all positively labeled sentences from an article
collectively gives the highest ROUGE score with respect to the gold
summary. ROUGE \cite{rouge}, a recall-oriented metric, is often used
to evaluate summarization systems. See Section~\ref{subsec:eval} for
more details. \newcite{nallapati17} reported comparable results
to \newcite{jp-acl16} with their automatically extracted labels on the
DailyMail dataset \cite{hermann-nips15}.

In our experiments we annotated the CNN dataset with
the \newcite{nallapati17} style annotation. We approach this
exponential problem of selecting the best subset of sentences using a
greedy approach and add one sentence at a time to the summary such
that the ROUGE score of the current summary is the highest with
respect to the gold summary. We stop adding new sentences to the
summary when the additions do not improve the ROUGE score or the
maximum number of sentences in the summary is reached.\footnote{We
choose maximum three sentences in the summary. See an explanation for
this in Section~\ref{subsec:eval}.}

We further augmented this dataset with side information. We used a
modified script of
\newcite{hermann-nips15} to extract titles and image captions, and we
associated them with the corresponding articles. All articles get
associated with their titles. The availability of image captions
varies from 0 to 414 per article, with an average of 3 image
captions. There are 40\% CNN articles with at least one image
caption. Our dataset is publicly available
at \url{https://github.com/shashiongithub/sidenet}.

We trained our network on a named-entity-anonymized version of news
articles.\footnote{We also experimented with the de-anonymized
articles, but the results were inferior to those presented here with
the anonymized data.} However, we generate de-anonymized summaries and
evaluate them against de-anonymized gold summaries to facilitate human
evaluation and to make human evaluation comparable to automatic
evaluation.

We used the standard splits of \newcite{hermann-nips15} for training,
validation and testing (90K/1,220/1,093 documents respectively).

\subsection{Comparison Systems}

We compared the output of our model against the standard baseline of
simply selecting the first three sentences from each document as the
summary. We refer to this baseline as \textsc{lead} in the rest of the
paper.

We also compared our system against the sentence extraction system of
\newcite{jp-acl16}.\footnote{The architecture
of \textsc{PointerNet} is closely related to the architecture
of \textsc{SideNet} without side information.} We refer to this system as
\textsc{PointerNet} as the neural attention architecture in
\newcite{jp-acl16} resembles the one in Pointer
Networks \cite{vinyals-nips15}. It does not exploit any side
  information.\footnote{Adding side information to \textsc{PointerNet}
  is an interesting direction of research but we do not pursue it
  here. It requires decoding with multiple types of attentions, and
  this is not the focus of this paper.}
%% but does have an attention  mechanism to attend to sentences while reading them.
 \newcite{jp-acl16} report only on the DailyMail dataset. We used
their code (\url{https://github.com/cheng6076/NeuralSum}) to produce
results on the CNN dataset.\footnote{We are unable to compare our
results to the extractive system of \newcite{nallapati17} because they
report their results on the DailyMail dataset and their code is not
available. The abstractive systems of \newcite{chenIjcai-16}
and \newcite{tanwan-acl17} report their results on the CNN dataset,
however, their results are not comparable to ours as they report on
the full-length F$_1$ variants of ROUGE to evaluate their abstractive
summaries. We report ROUGE recall scores which is more appropriate to
evaluate our extractive summaries.}

%% As the results of their system are not available on the CNN
%% dataset, we have implemented \textsc{PointerNet} in
%% TensorFlow.\footnote{For sanity check, we have trained our
%% implementation of \textsc{PointerNet} on the DailyMail dataset and
%% achieved comparable results to what has been reported
%% by \newcite{jp-acl16}.}

%% The abstractive systems of \newcite{chenIjcai-16}
%% and \newcite{tanwan-acl17} also report result on the CNN dataset,
%% however, their results are not comparable to ours as they generate
%% named-entity anonymized summaries.

%% We are grateful to the authors of \textsc{PointerNet}
%% for providing us with the outputs of their system on both CNN and
%% DailyMail datasets. We could not produce results for
%% DailyMail-Filtered, as \textsc{PointerNet} uses \textsc{SentExtLabels}
%% and it is not available for DailyMail-Filtered.

%%  To make this comparison possible, we implemented their system from
%% scratch in TensorFlow with few variations. For examples, our
%% implementation of \textsc{PointerNet} is not trained with the
%% curriculum learning strategy of
%% \newcite{bengio-nips2015-curriculum} to reduce the exposure bias
%% \cite{ranzato-arxiv15-bias}.

\subsection{Implementation Details}

We used our training data to train word embeddings using the
\textit{Word2vec} skip-gram model \cite{word2vec} with context window
size 6, negative sampling size 10 and hierarchical softmax 1.  For
known words, word embedding variables were initialized with
pre-trained word embeddings of size 200. For unknown words, embeddings
were initialized to zero, but optimized during training. All
sentences, including titles and image captions, were padded with zeros
to a sentence length of 100. For the convolutional sentence encoder,
we followed
\newcite{kim-aaai16}, and used a list of kernels of widths 1 to 7,
each with output channel size of 50. This leads the sentence embedding
size in our model to be 350. For the recurrent neural network
component in document encoder and sentence extractor, we used a
single-layered LSTM network with size 600. All input documents were
padded with zeros to a maximum document length of 126. For each
document, we consider a maximum of 10 image captions. We experimented
with various numbers (1, 3, 5, 10 and 20) of image captions on the
validation set and found that our model performed best with 10 image
captions. We performed mini-batch cross-entropy training with a batch
size of 20 documents for 10 training epochs. After each epoch, we
evaluated our model on the validation set and chose the best
performing model for the test set. We trained our models with the
optimizer Adam \cite{adam-14} with initial learning rate 0.001. Our
system is fully implemented in
TensorFlow \cite{tensorflow2015-whitepaper}.\footnote{Our TensorFlow
code is publicly available at \url{https://github.com/shashiongithub/sidenet}.}

\section{Results and Discussion}
\label{sec:results}

We conducted an automatic and a human evaluation. We start this
section with an ablation study on the validation set. The best model
from this study is chosen for the test set. In the rest of the paper,
we refer to our model as \textsc{SideNet} for its ability to exploit
side information.

\begin{table}[t!]
  \center{\footnotesize
  \begin{tabular}{ l | c c c c c r }
    \hline 
    \textsc{Models} & R1 & R2 & R3 & R4 & RL &  Avg.\\ \hline
    \textsc{lead} & 49.2 & 18.9 & 9.8 & 6.0 & 43.8 & 25.5   \\ 
    \textsc{PointerNet} & 53.3 & 19.7 & 10.4 & 6.4 & 47.2 & 27.4 \\ 
    % \textsc{SideNet-} & 54.3 & 21.1 & 11.1 & 6.9 & 48.9 & 28.5 \\ 
    \hline \hline
    \textsc{SideNet+title} & 55.0 & 21.6 & 11.7 & 7.5 & 48.9 & 28.9  \\
    \textsc{SideNet+caption} & 55.3 & 21.3 & 11.4 & 7.2 & 49.0 & 28.8   \\
    \textsc{SideNet+fs} & 54.8 & 21.1 & 11.3 & 7.2 & 48.6 & 28.6\\ \hline
    \multicolumn{7}{c}{Combination Models (\textsc{SideNet+})} \\ \hline 
    \textsc{title+caption} & \textbf{55.4} & \textbf{21.8} & \textbf{11.8} & \textbf{7.5} & \textbf{49.2} & \textbf{29.2}  \\
    \textsc{title+fs} & 55.1 & 21.6 & 11.6 & 7.4 & 48.9 & 28.9 \\
    \textsc{caption+fs} & 55.3 & 21.5 & 11.5 & 7.3 & 49.0 & 28.9  \\
    \textsc{title+caption+fs} & 55.4 & 21.5 & 11.6 & 7.4 & 49.1 & 29.0 \\
    \hline
  \end{tabular}
  }

  \caption{Ablation results on the validation set. We report R1, R2,
    R3, R4, RL and their average (Avg.). The first block of the table
    presents \textsc{lead} and \textsc{PointerNet} which do not use
    any side information. \textsc{lead} is the baseline system
    selecting ``first'' three sentences. \textsc{PointerNet} is the
    sentence extraction system of Cheng and
    Lapata\nocite{jp-acl16}. \textsc{SideNet} is our model. The second
    and third blocks of the table present different variants
    of \textsc{SideNet}. We experimented with three types of side
    information: title (\textsc{title}), image captions
    (\textsc{caption}) and the first sentence (\textsc{fs}) of the
    document. The bottom block of the table presents models with more
    than one type of side information. The best performing model
    (highlighted in boldface) is used on the test
    set.} \label{tab:cnnablation}
\end{table}

\begin{table}[t!]
  \center{\footnotesize % \scriptsize %\footnotesize
  \begin{tabular}{ l | c c c c r } 
    \hline
    \textsc{Models} & \textsc{R1} & \textsc{R2} & \textsc{R3} & \textsc{R4} & \textsc{RL} \\ \hline
    \multicolumn{6}{c}{Fixed length: 75b}\\
    \hline 
    \textsc{lead} & 20.1 & 7.1 & \textbf{3.5} & 2.1 & 14.6 \\
    \textsc{PointerNet} &  \textbf{20.3} & \textbf{7.2} & \textbf{3.5} & \textbf{2.2} & \textbf{14.8} \\
    \textsc{SideNet} & 20.2 & 7.1 & 3.4 & 2.0 & 14.6 \\
    \hline
      \multicolumn{6}{c}{Fixed length: 275b}\\ 
    \hline
    \textsc{lead} & 39.1 & 14.5 & 7.6 & 4.7 & 34.6 \\
    \textsc{PointerNet} & 38.6 & 13.9 & 7.3 & 4.4 & 34.3 \\
    \textsc{SideNet} & \textbf{39.7} & \textbf{14.7} & \textbf{7.9} & \textbf{5.0} & \textbf{35.2} \\
    \hline
     \multicolumn{6}{c}{Full length summaries}\\ 
    \hline
    \textsc{lead} & 49.3 & 19.5 & 10.7 & 6.9 & 43.8 \\
    \textsc{PointerNet} &  51.7 & 19.7 & 10.6 & 6.6 & 45.7 \\
    \textsc{SideNet} & \textbf{54.2} & \textbf{21.6} & \textbf{12.0} & \textbf{7.9} & \textbf{48.1} \\
    \hline
  \end{tabular}
  }
  \caption{Final results on the test set. \textsc{PointerNet}
    is the sentence extraction system of Cheng and
    Lapata\nocite{jp-acl16}. \textsc{SideNet} is our best model from
    Table \ref{tab:cnnablation}. Best ROUGE score in each block and
    each column is highlighted in boldface.}
  \label{tab:rougerefinal}
\end{table}

\subsection{Automatic Evaluation}
\label{subsec:eval}

To automatically assess the quality of our summaries, we used ROUGE
\cite{rouge}, a recall-oriented metric, to compare our model-generated
summaries to manually-written highlights.\footnote{We used pyrouge, a
  Python package, to compute all our rouge scores with parameters ``-a
  -c 95 -m -n 4 -w 1.2.''} Previous work has reported ROUGE-1 (R1) and
ROUGE-2 (R2) scores to access informativeness, and ROUGE-L (RL) to
access fluency. In addition to R1, R2 and RL, we also report ROUGE-3
(R3) and ROUGE-4 (R4) capturing higher order $n$-grams overlap to
assess informativeness and fluency simultaneously.

We follow Cheng and Lapata and report on both full length (three
sentences with the top scores as the summary) and fixed length (first
75 bytes and 275 bytes as the summary) summaries. For full length
summaries, our decision of selecting three sentences is guided by the
fact that there are 3.11 sentences on average in the gold highlights
of the training set.  We conduct our ablation study on the validation
set with full length ROUGE scores, but we report both fixed and full
length ROUGE scores for the test set.

We experimented with two types of side information: title
(\textsc{title}) and image captions (\textsc{caption}). In addition,
we experimented with the first sentence (\textsc{fs}) of the document
as side information. Note that the latter is not strictly speaking
side information, it is a sentence in the document. However, we wanted
to explore the idea that the first sentence of the document plays a
crucial part in generating summaries
\cite{rush-acl15,nallapati-signll16}. \textsc{SideNet} with \textsc{fs} acts as a baseline
for \textsc{SideNet} with title and image captions.

%% We also experimented with the document length of the article. We
%% found that our model performed best for the document length of 10
%% sentences for CNN articles and 50 for DailyMail articles. This
%% could be due to the differences in writing styles of articles in a
%% main-stream news network (CNN) vs a tabloid newspaper
%% (DailyMail).\footnote{These descriptions of CNN and DailyMail are
%% taken from Wikipedia.} All parameters were tuned on validation
%% sets.

We report the performance levels of several variants of
\textsc{SideNet} on the validation set in Table
\ref{tab:cnnablation}. We also compare them against the \textsc{lead}
baseline and \textsc{PointerNet}. These two systems do
not use any side information. Interestingly, all the variants
of \textsc{SideNet} significantly outperform \textsc{lead}
and \textsc{PointerNet}.  When the title (\textsc{title}), image
captions (\textsc{caption}) and the first sentence (\textsc{fs}) are
used separately as side information, \textsc{SideNet} performs best
with \textsc{title} as its side information. Our result demonstrates
the importance of the title of the document in extractive
summarization
\cite{Edmundson:1969,Kupiec:1995binary,mani2001automatic}. 
The performance with \textsc{title} and \textsc{caption} is better
than that with \textsc{fs}. We also tried possible combinations
of \textsc{title}, \textsc{caption} and \textsc{fs}. All
\textsc{SideNet} models are superior to the ones without any side information.
\textsc{SideNet} performs best when \textsc{title} and \textsc{caption} are
jointly used as side information (55.4\%, 21.8\%, 11.8\%, 7.5\%, and
49.2\% for R1, R2, R3, R4, and RL respectively). It is better than the
the \textsc{lead} baseline by 3.7 points on average and
than \textsc{PointerNet} by 1.8 points on average, indicating that
side information is useful to identify the gist of the document. We
use this model for testing purposes.

Our final results on the test set are shown in Table
\ref{tab:rougerefinal}. We present both fixed length (first 75 bytes
and 275 bytes) and full length (three highest scoring sentences) ROUGE
scores. It turns out that for smaller summaries (75 bytes)
\textsc{lead} and \textsc{PointerNet} are superior to
\textsc{SideNet}. This result could be because \textsc{lead} (always)
and \textsc{PointerNet} (often) include the first sentence in their
summaries, whereas, \textsc{SideNet} is better capable at selecting
sentences from various document positions. This is not captured by
smaller summaries of 75 bytes, but it becomes more evident with longer
summaries (275 bytes and full length) where \textsc{SideNet} performs
best across all ROUGE scores. It is interesting to note that
\textsc{PointerNet} performs better than \textsc{lead} for 75-byte
summaries, then its performance drops behind \textsc{lead} for
275-byte summaries, but then it performs better than \textsc{lead} for
full length summaries for R1, R2 and RL. It shows that
\textsc{PointerNet} with its attention over sentences in the document
is capable of exploring more than first few sentences in the
document. But, it is still behind \textsc{SideNet} which is better at
identifying salient sentences in the document. \textsc{SideNet}
performs better than \textsc{PointerNet} by 0.8 points for 275-byte
summaries and by 1.9 points for full length summaries, on average for
all ROUGE scores.

\begin{table}[t!]
  \center{\footnotesize
    \begin{tabular}{ l | c c c r } 
      \hline
      Models & 1st & 2nd & 3rd & 4th \\ \hline
      \textsc{lead} & 0.15 & 0.17 & \textbf{0.47} & 0.21 \\ 
      \textsc{PointerNet} & 0.16 & 0.05 & 0.31 & \textbf{0.48} \\
      \textsc{SideNet} & 0.28 & \textbf{0.53} & 0.15 & 0.04 \\ 
      \textsc{human} & \textbf{0.41} & 0.25 & 0.07 & 0.27 \\ 
      \hline
    \end{tabular}}
  \caption{Human evaluations: Ranking of various systems. Rank
    1st is best and rank 4th, worst. Numbers show the percentage of
    times a system gets ranked at a certain
    position.} \label{tab:heval}
\end{table}

\begin{figure}[t!]
  \center{ % \small % footnotesize % scriptsize % footnotesize
    \begin{tabular}{| c | p{7.4cm} |}
      \hline 
      \multirow{8}{*}{\rotatebox[origin=c]{90}{\textsc{lead}}} & 
      \textbullet \hspace{0.1cm} Seoul South korea's Prime Minister
      Lee Wan-koo offered to resign on monday amid a growing political
      scandal \\
      & \textbullet \hspace{0.1cm} Lee will stay in his official role
      until South Korean President Park Geun-hye accepts his
      resignation \\
      & \textbullet \hspace{0.1cm} He has transferred his role of
      chairing cabinet meetings to the deputy Prime Minister for the
      time being , according to his office \\ \hline

      \multirow{7}{*}{\rotatebox[origin=c]{90}{\textsc{PointerNet}}} &
      \textbullet \hspace{0.1cm} South Korea's Prime Minister Lee
      Wan-koo offered to resign on Monday amid a growing political
      scandal \\
      & \textbullet \hspace{0.1cm} Lee will stay in his official role
      until South Korean President Park Geun-hye accepts his
      resignation \\
      & \textbullet \hspace{0.1cm} Lee and seven other politicians
      with links to the South Korean President are under
      investigation \\ \hline
   
      \multirow{8}{*}{\rotatebox[origin=c]{90}{\textsc{SideNet}}} &
      \textbullet \hspace{0.1cm} South Korea's Prime Minister Lee
      Wan-Koo offered to resign on Monday amid a growing political
      scandal \\
      & \textbullet \hspace{0.1cm} Lee will stay in his official role
      until South Korean President Park Geun-hye accepts his
      resignation \\
      & \textbullet \hspace{0.1cm} Calls for Lee to resign began after
      South Korean tycoon Sung Woan-jong was found hanging from a
      tree in Seoul in an apparent suicide on April 9 \\ \hline
      
      \multirow{6}{*}{\rotatebox[origin=c]{90}{\textsc{human}}} &
      \textbullet \hspace{0.1cm} Calls for Lee Wan-koo to resign
      began after South Korean tycoon Sung Woan-jong was found hanging
      from a tree in Seoul \\
      & \textbullet \hspace{0.1cm} Sung, who was under investigation for
      fraud and bribery, left a note listing names and amounts of cash
      given to top officials \\ \hline
    \end{tabular}
  }
  \caption{Summaries produced by various systems for the
    article shown in Figure \ref{fig:cnnex}.}\label{fig:summaries}
\end{figure}

\subsection{Human Evaluation}

We complement our automatic evaluation results with human
evaluation. We randomly selected 20 articles from the test set.
Annotators were presented with a news article and summaries from four
different systems. These include the \textsc{lead} baseline,
\textsc{PointerNet}, \textsc{SideNet} and the human authored
highlights. We followed the guidelines in \newcite{jp-acl16}, and
asked our participants to rank the summaries from best (1st) to worst
(4th) in order of informativeness (does the summary capture important
information in the article?) and fluency (is the summary written in
well-formed English?). We did not allow any ties, we only sampled
articles with non-identical summaries. We assigned this task to five
annotators who were proficient English speakers. Each annotator was
presented with all 20 articles. The order of summaries to rank was
randomized per article. Examples of summaries our subjects ranked are
shown in Figure \ref{fig:summaries}.

%% As a result, we collected 5 responses per article ranking each
%% summary 100 times.

The results of our human evaluation study are shown in Table
\ref{tab:heval}. We compare our \textsc{SideNet} against
\textsc{lead}, \textsc{PointerNet} and \textsc{human} on how
frequently each system gets ranked 1st, 2nd and so on, in terms of
best-to-worst summaries. As one might imagine, \textsc{human} gets
ranked 1st most of the time (41\%). However, it is closely followed by
\textsc{SideNet} with ranked 1st 28\% of the time. In comparison,
\textsc{PointerNet} and \textsc{lead} were mostly ranked at 3rd and
4th places. We also carried out pairwise comparisons between all
models in Table \ref{tab:heval} for their statistical significance
using a one-way ANOVA with post-hoc Tukey HSD tests with (p $<$
0.01). It showed that \textsc{SideNet} is significantly %% (p$=$0.001)
better than \textsc{lead} and \textsc{PointerNet}, and it does not
differ significantly %% (p$=$0.33)
from \textsc{human}. On the other
hand, \textsc{PointerNet} does not differ significantly % (p$=$0.10)
from \textsc{lead} and it differs significantly from both
\textsc{SideNet} and \textsc{human}. %% (for both p$=$0.001)
The human evaluation results corroborates our empirical results in
Table \ref{tab:cnnablation} and Table \ref{tab:rougerefinal}:
\textsc{SideNet} is better than \textsc{lead} and \textsc{PointerNet}
in producing informative and fluent summaries.

%% A vs B  3.2449  0.1014344	insignificant
%% A vs C  7.0306  0.0010053	** p<0.01
%% A vs D  4.6510  0.0061261	** p<0.01
%% B vs C  10.2755 0.0010053	** p<0.01
%% B vs D  7.8959  0.0010053	** p<0.01
%% C vs D  2.3796  0.3347295	insignificant

Figure \ref{fig:summaries} shows output summaries from various systems
for the article shown in Figure \ref{fig:cnnex}. As can be seen,
both \textsc{SideNet} and \textsc{PointerNet} were able to select the
most relevant sentence for the summary from anywhere in the article,
but \textsc{SideNet} is better at producing summaries which are close
to human authored summaries.

\section{Conclusion}
\label{sec:conclusion}

In this paper, we developed a neural network framework for
single-document extractive summarization with side information. We
evaluated our system on the large scale CNN dataset. Our experiments
show that side information is useful for extracting salient sentences
from the document for the summary. Our framework is very general and
it could exploit different types of side information.  There are few
previous works which improve extractive summarization with external
knowledge from third party sources. \newcite{svore-emnlp07} included
features from news search query logs and Wikipedia entities to
summarize CNN articles. Recently,
\newcite{li-coling16} used public posts following a news article to
improve automatic summarization. For future work, it would be interesting to use such
knowledge as side information in our framework.

\section*{Acknowledgments} 

We thank Jianpeng Cheng for providing us with the CNN
dataset and an implementation of PointerNet. We also thank Laura Perez
and members of the ILCC Cohort group for participating in our
human evaluation experiments. This work greatly benefitted from
discussions with Jianpeng Cheng, Annie Louis, Pedro Balage, Alfonso
Mendes, Sebasti\~{a}o Miranda, and members of the ILCC ProbModels
group. This research is supported by the H2020 project SUMMA (under grant
agreement 688139).

%References and End of Paper
%These lines must be placed at the end of your paper
\bibliography{summarisation-improved-arxiv}
\bibliographystyle{aaai}
\end{document}